\documentclass{article}
\usepackage{spconf}
\usepackage{amsmath,amssymb}
\usepackage{graphicx}
\usepackage{enumitem}
\usepackage{titlesec}
\usepackage{url}

\def\x{\mathbf{x}}          % input image
\def\z{\mathbf{z}}          % latent vector
\def\p{\mathbf{p}}          % textual prompt
\def\A{\mathcal{A}}         % set of annotator masks
\def\L{\mathcal{L}}         % loss

\title{ProSona: Prompt-Guided Personalization for Multi-Expert Medical Image Segmentation}

\name{
\begin{tabular}{c}
Aya~Elgebaly$^{8}$\thanks{A. Elgebaly has done the work during her Master's internship at Albarqouni Lab. at the University Hospital Bonn, Germany.} \quad
Nikolaos~Delopoulos$^{2}$ \quad
Juliane~Hörner\mbox{-}Rieber$^{3,4}$ \quad
Carolin~Rippke$^{4}$ \\
\textit{Sebastian~Klüter$^{4}$ \quad
Luca~Boldrini$^{5}$ \quad
Lorenzo~Placidi$^{5}$ \quad
Riccardo~Dal~Bello$^{6}$ \quad
Nicolaus~Andratschke$^{6}$}\\
\textit{Michael~Baumgartl$^{6}$ \quad
Claus~Belka$^{2,7}$ \quad
Christopher~Kurz$^{2}$ \quad
Guillaume~Landry$^{2,7}$}\\
\textit{Shadi~Albarqouni$^{1,9}$}\thanks{S.Albarqouni is the corresponding author: shadi.albarqouni@ukbonn.de}
\end{tabular}
}

\address{
$^{1}$Clinic for Diagnostic and Interventional Radiology, University Hospital Bonn, Germany \\
$^{2}$Department of Radiation Oncology, LMU University Hospital, LMU Munich, Germany \\
$^{3}$Department of Radiation Oncology, University Hospital Düsseldorf, Germany \\
$^{4}$Department of Radiation Oncology, Heidelberg University Hospital, Germany \\
$^{5}$Università Cattolica del Sacro Cuore, Rome, Italy \\
$^{6}$Department of Radiation Oncology, University Hospital Zurich, Switzerland\\
$^{7}$Bavarian Cancer Research Center (BZKF), Munich, Germany\\
$^{8}$ Department of Applied Mathematics and Computer Science, DTU Compute, Denmark\\
$^{9}$ Helmholtz AI, Helmholtz Center Munich, Germany
}

\begin{document}
\ninept 
\maketitle
%

% -------------------------------------------------------------------------
\begin{abstract}
Automated medical image segmentation suffers from high inter-observer variability, particularly in tasks such as lung nodule delineation, where experts often disagree.  
Existing approaches either collapse this variability into a consensus mask or rely on separate model branches for each annotator.  
We introduce \verb|ProSona|, a two-stage framework that learns a continuous latent space of annotation styles, enabling controllable personalization via natural language prompts.  
A probabilistic U-Net backbone captures diverse expert hypotheses, while a prompt-guided projection mechanism navigates this latent space to generate personalized segmentations.  
A multi-level contrastive objective aligns textual and visual representations, promoting disentangled and interpretable expert styles.  
Across the LIDC–IDRI lung nodule and multi-institutional prostate MRI datasets, \verb|ProSona| reduces the Generalized Energy Distance by 17\% and improves mean Dice by more than one point compared with DPersona.  
These results demonstrate that natural-language prompts can provide flexible, accurate, and interpretable control over personalized medical image segmentation. Our implementation is available online\footnote{\url{https://github.com/albarqounilab/ProSona}}.
\end{abstract}

% List of keywords.

%Introduction 
\section{Introduction} 
\label{sec:intro}
Medical image segmentation underpins numerous clinical tasks, from tumor delineation to surgical and radiotherapy planning, yet object boundaries are ambiguous. Two sources of uncertainty complicate this process: \textit{data-level ambiguity}, arising from poor contrast, partial volume effects, or irregular morphologies, and \textit{observer-level variability}, driven by differences in expert judgment and institutional guidelines. In datasets such as LIDC--IDRI, radiologists frequently disagree on lung nodule boundaries~\cite{armato2011lung,joskowicz2019inter}, while similar inconsistencies appear in prostate MRI segmentation across institutions.

\begin{figure}[t]
\centering
\includegraphics[width=\linewidth]{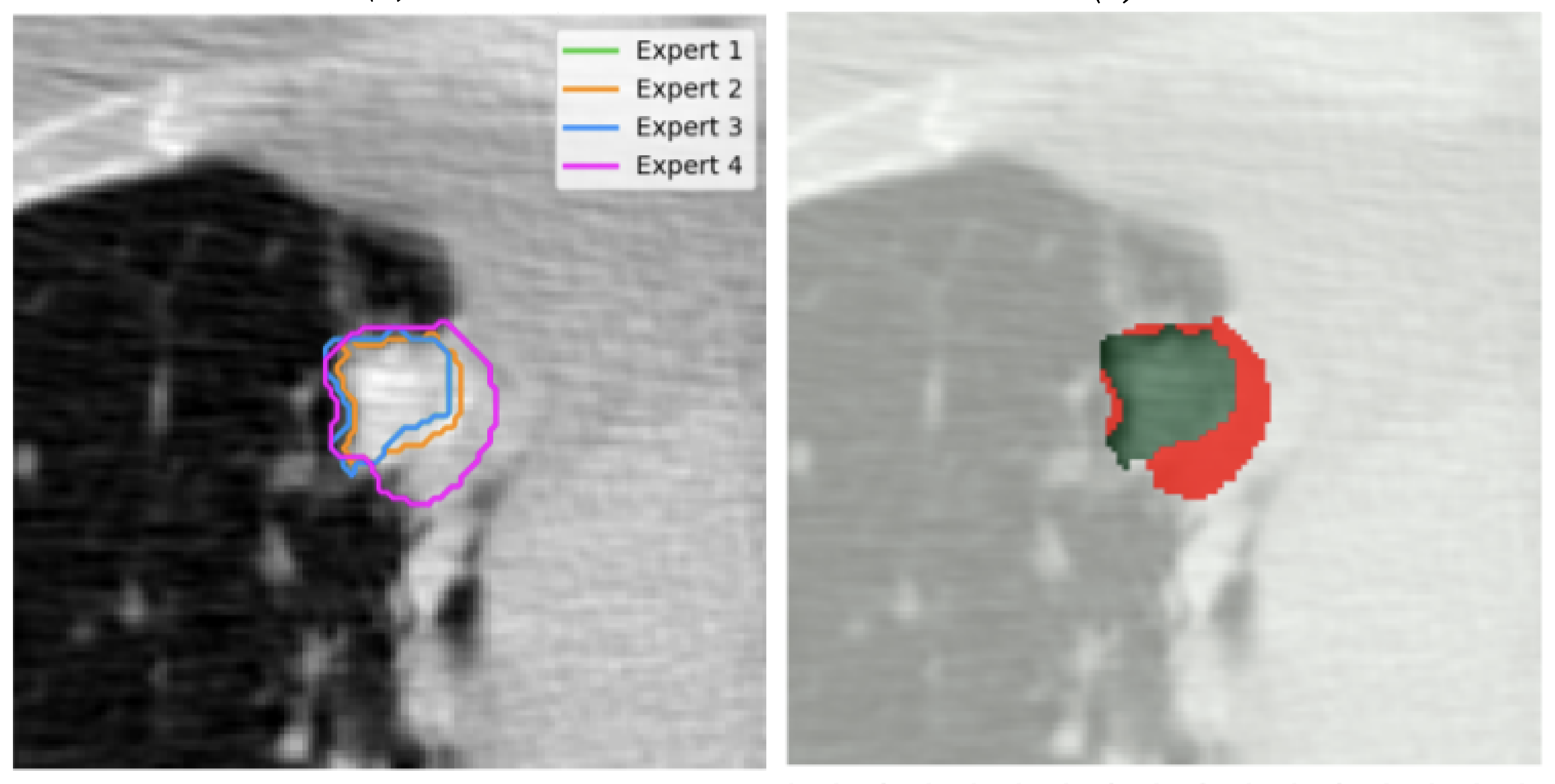}
\caption{Information loss in consensus fusion for lung nodule. Left: expert delineations showing variability in boundary interpretation.  
Right: majority-vote mask (green) and regions omitted by consensus (red), corresponding to subtle infiltrative extensions potentially relevant for treatment planning.}
\label{fig:mv}
\end{figure}

%These discrepancies are not trivial. 
Consensus-based fusion, such as majority voting, can suppress subtle but clinically significant regions, for example, infiltrative tumor margins or peritumoral extensions that indicate microscopic invasion (Fig.~\ref{fig:mv}). Such omissions may underestimate tumor aggressiveness or exclude disease regions during treatment planning. Capturing this diversity is therefore crucial for building models that reflect the full spectrum of plausible expert interpretations.

Multi-rater segmentation has evolved along three main lines. Early label-fusion algorithms such as STAPLE~\cite{warfield2004simultaneous} yield a single consensus mask but discard inter-annotator variability. Generative models, including the Probabilistic U-Net and PHiSeg~\cite{kohl2018probabilistic,baumgartner2019phiseg}, learn a latent distribution over plausible masks to sample diverse hypotheses. MoDiff~\cite{modiff2025} replaces one‑hot masks with probability‑based labels and employs a morphology‑emphasized diffusion model to capture high‑frequency boundaries better. Multi‑rater prompting uses lightweight rater‑aware prompts to handle ambiguous segmentation with minimal parameter updates~\cite{wang2024multirater}.
Recent personalization-based methods aim to model annotator-specific outputs, either through multiple decoder heads or shared latent spaces. DPersona~\cite{wu2024diversified} exemplifies this approach but suffers from architectural inefficiency and weak style disentanglement due to image-dependent style selection.

To address these challenges, we propose \verb|ProSona|, a prompt-guided personalization framework that models expert diversity within a continuous, semantically navigable latent space controllable via natural-language prompts.  
\verb|ProSona| produces multi-hypothesis segmentations that preserve annotator variability and allow interpretable exploration of expert styles.
Our contributions:
\begin{enumerate}[leftmargin=14pt, nosep]
  \item A \textbf{prompt-guided personalization module} that maps language descriptions of annotators into a shared latent space, generating similarity-weighted latent codes for personalized predictions.
  \item A \textbf{multi-level contrastive objective} that aligns textual and visual representations, improving style disentanglement and enabling smooth interpolation between annotators.
  \item Extensive evaluation on the \textbf{LIDC--IDRI lung nodule dataset} and a \textbf{multi-institutional in-house prostate MRI dataset}, showing superior fidelity to individual annotators/styles.
\end{enumerate}

% Method 
\section{Proposed Method}
\label{sec:methods}

Given an image $\x \in \mathbb{R}^{H\times W\times C}$ and its set of expert masks $\A=\{a^{(1)},\dots,a^{(A)}\}$ with each $a^{(i)}\in\{0,1\}^{H\times W}$, our goal is to learn a function 
$\hat{y} =f(\x,\p)\!\rightarrow\!\{0,1\}^{H\times W}$ that generates a segmentation consistent with a textual prompt $\p$ describing an annotator’s style (e.g., ``conservative segmentation'' or ``inclusive segmentation'').

\begin{figure*}[t]
\centering
\includegraphics[width=\linewidth]{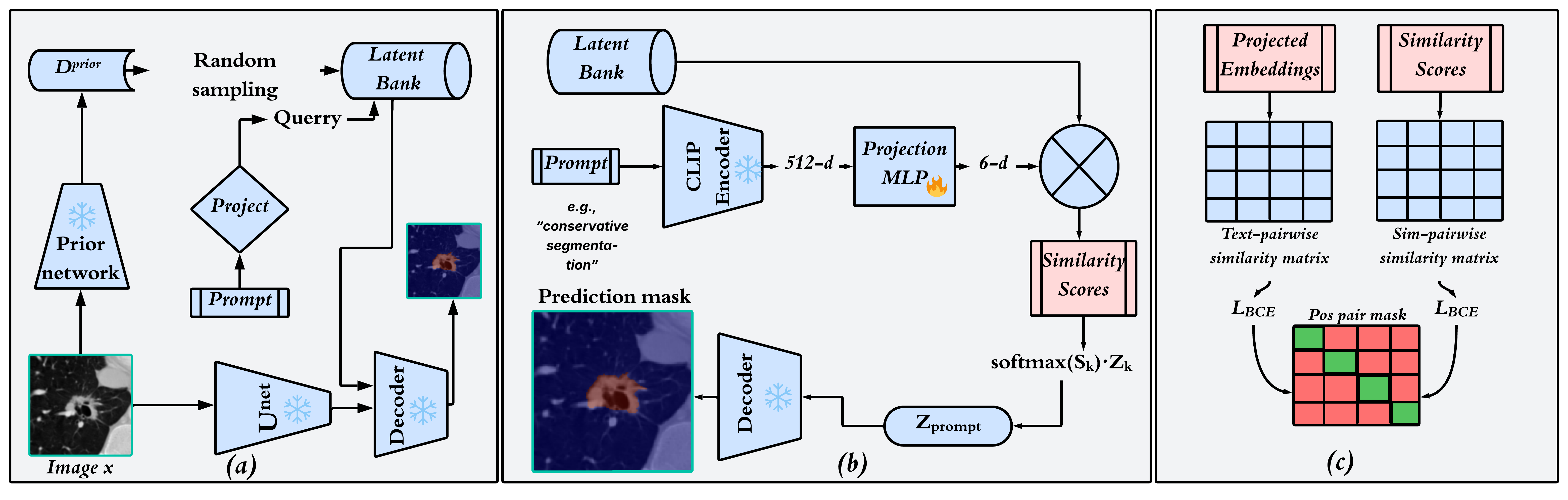}
\caption{Overview of our ProSona: (a) Stage 2: Personalization - enabling text-guided segmentation through the prior bank and prompt-based latent space navigation; (c) Prompt processing pipeline showing CLIP encoding and similarity-based latent selection; (d) Multi-level contrastive learning with text-pairwise and similarity-pairwise matrices for better disentanglement of annotation styles.}
\label{fig:framework_overview}
\end{figure*}

\paragraph{Method Overview.}
\verb|ProSona| comprises two sequential stages (Fig.~\ref{fig:framework_overview}). 
Stage~1 constructs a \emph{multi-hypothesis latent space} that explicitly models the distribution of expert annotations, overcoming the collapse of inter-annotator variability inherent in consensus or deterministic methods. 
Stage~2 introduces \emph{prompt-guided personalization}, where language descriptions of annotators are projected into this latent space to retrieve or interpolate between expert styles, offering intuitive and interpretable control. 
Finally, a dual contrastive objective aligns textual and latent representations, enforcing consistent behavior across images and preventing style entanglement. 
Together, these components yield a semantically navigable latent space enabling controllable, interpretable, and clinically faithful multi-hypothesis predictions.

\subsection{Stage 1: Latent Space Construction}
We follow DPersona \cite{wu2024diversified} by first building a diverse latent space capturing annotator variability, addressing the limitation of single-distribution models that average out expert disagreement.
Following the Probabilistic U-Net~\cite{kohl2018probabilistic}, the encoder produces a deterministic feature map and two convolutional heads estimate parameters $(\boldsymbol\mu,\boldsymbol\sigma)$ of a Gaussian prior $\mathcal{N}(\boldsymbol\mu,\operatorname{diag}(\boldsymbol\sigma))$. 
A posterior network conditioned on both $\x$ and a randomly selected annotation $a$ yields $\mathcal{N}(\boldsymbol\mu',\operatorname{diag}(\boldsymbol\sigma'))$. 
A latent code $\z$ sampled from the posterior is concatenated with encoder features before decoding to a segmentation $\hat{y}$.

The Stage~1 objective encourages accurate reconstruction, distributional alignment, and boundary diversity:
\[
\L_{\mathrm{stage1}} = \L_{\mathrm{seg}} + \L_{\mathrm{KL}} + \L_{\mathrm{bound}},
\]
where $\L_{\mathrm{seg}}$ is a Dice loss between $\hat{y}$ and $a$, and $\L_{\mathrm{KL}}$ is the Kullback–Leibler divergence between posterior and prior.  
$\L_{\mathrm{bound}}$ enforces that the ensemble of sampled segmentations matches the experts’ intersection and union, preventing mode collapse and promoting boundary variability.  
Let $\{\hat{y}_1,\!\dots,\!\hat{y}_K\}$ denote predictions from $K$ latent samples. 
The predicted intersection and union are $P_{\cap}$ and $P_{\cup}$, giving
\[
\L_{\mathrm{bound}} = \operatorname{Dice}(P_{\cap},A_{\cap}) + \operatorname{Dice}(P_{\cup},A_{\cup}),
\]
where $A_{\cap}$ and $A_{\cup}$ are the intersection and union of expert masks in $\A$.
Intuitively, this stage teaches the model to generate plausible annotation variants before learning how prompts select among them.

\subsection{Stage 2: Prompt-Guided Personalization}
While Stage~1 captures the diversity of expert annotations, Stage~2 (Fig.~\ref{fig:framework_overview}, b) enables explicit and interpretable control within this space via natural-language prompts, eliminating the need for annotator-specific branches, as appeared in DPersona \cite{wu2024diversified}, and improving scalability.
A pre-trained, frozen CLIP text encoder embeds the prompt $\p$ into a feature vector $e\in\mathbb{R}^d$, which an MLP projects into the latent space \(\mathbb{R}^D\), producing $\z_{\mathrm{proj}}\in\mathbb{R}^D$.  
To locate the relevant expert region, we draw $K$ latent samples $\{\z_k\}_{k=1}^K\in\mathbb{R}^D$ from the prior and compute similarity scores 
$s_k = (\z_{\mathrm{proj}}^{\top}\!\z_k)/\sqrt{D}$, where $D$ is the latent dimension.  
The prompt-specific latent code is a softmax-weighted combination:
\[
\z_{\mathrm{prompt}} = \sum_{k=1}^K \operatorname{softmax}(s_k)\,\z_k,
\]

This vector is concatenated with deterministic U-Net features and passed to the decoder to generate a personalized prediction $\hat{y}_{\mathrm{prompt}}$.
This mechanism allows smooth prompt-conditioned interpolation between annotator styles, enabling generation of new plausible annotation hypotheses and intuitive model steering.

\subsection{Multi-Level Contrastive Disentanglement}
To ensure that prompt–latent correspondence learned in Stage~2 remains consistent and interpretable, we introduce two complementary contrastive objectives (Fig.~\ref{fig:framework_overview}, c).  
These objectives align linguistic and visual representations of annotators across images, strengthening style consistency and disentanglement.  
Let $E\!\in\!\mathbb{R}^{P\times d}$ denote normalized prompt embeddings (for \(P\) annotator prompts) and $R\!\in\!\mathbb{R}^{P\times K}$ the corresponding normalized similarity profiles (similarities between each prompt and $K$ latent samples).  
A binary mask $M\!\in\!\{0,1\}^{P\times P}$ encodes positive pairs ($M_{ij}=1$ for prompts from the same annotator).  
We define
\[
\L_{\mathrm{text}} = \L_{\mathrm{BCE}}\!\big((E E^\top)/\tau,\,M\big),
\L_{\mathrm{sim}} = \L_{\mathrm{BCE}}\!\big((R R^\top)/\tau,\,M\big),
\]
where $\tau$ is a temperature parameter.  
The total Stage~2 loss combines segmentation and contrastive terms:
\[
\L_{\mathrm{stage2}} = \L_{\mathrm{seg}} + \alpha\,\L_{\mathrm{text}} + \beta\,\L_{\mathrm{sim}},
\]
with $\alpha$ and $\beta$ balancing the contrastive objectives.  
Together, these losses promote consistent alignment between textual and latent spaces, prevent mode collapse, and yield a semantically navigable expert manifold that supports prompt-based multi-hypothesis generation.

% Experiments 
\section{Experiments}
\label{sec:experiments}

The experiments aim to assess whether \verb|ProSona| can (i) accurately capture inter-annotator variability through a semantically navigable latent space and (ii) generalize its prompt-guided personalization across institutions.

\paragraph{Datasets.}
We evaluate on two settings: (1) the public \textbf{LIDC–IDRI} lung nodule dataset~\cite{armato2011lung}, containing 1{,}609 two-dimensional CT slices each annotated by four radiologists, and (2) an \textbf{in-house multi-institutional prostate MRI} dataset collected from four clinical centres (LMU University Hospital, Gemelli University Hospital, University Hospital Zurich, and Heidelberg University Hospital).  
For LIDC–IDRI, we follow~\cite{wu2024diversified} in ordering annotators from conservative to inclusive based on mask area.  
For prostate MRI, each slice is labelled by a single expert; to emulate multi-rater variability while preserving privacy, we generate three additional pseudo-masks per image using independent nnU-Net models trained on each institution’s data.  
All images are resampled to isotropic spacing, intensity-normalized, and cropped to $128\times128$.  
The final dataset comprises 4{,}817 slices.  
Four-fold cross-validation is performed at the patient level for both.

\paragraph{Implementation Details.}
We adopt DPersona’s U-Net backbone with a six-dimensional latent code.  
Stage~1 and Stage~2 are trained for 100 epochs each using Adam with a learning rate $10^{-4}$ and a batch size of eight.  
We use $K{=}10$ latent samples per image and a frozen CLIP text encoder to embed annotator descriptions (e.g., ``conservative mask,'' ``inclusive mask'').  
Prompts are projected into the latent space via a two-layer MLP.  
The prostate MRI experiments use an identical architecture and hyperparameters as LIDC–IDRI.

\paragraph{Evaluation Metrics.}
We compare against single-annotator U-Nets, CM-Global~\cite{tanno2019learning}, CM-Pixel~\cite{zhang2020disentangling}, TAB~\cite{liao2023transformer}, Pionono~\cite{schmidt2023probabilistic}, and DPersona~\cite{wu2024diversified}.  
Performance is reported using \textit{Generalized Energy Distance (GED)}, \textit{Dice Soft} (probabilistic Dice averaged over samples), \textit{Dice Max} (best prediction–ground truth match), \textit{Dice Match} (Dice with the corresponding annotator), and the \textit{Mean Dice} across annotators or institutions.

% -------------------------------------------------------------------------

\paragraph{Quantitative Performance.}

\begin{table}[t]
\centering
\caption{Comparison with baseline methods on LIDC–IDRI (upper block) and prostate MRI (lower block).  We report GED (\textdownarrow), Dice Soft (\textuparrow), Dice Max (\textuparrow), Dice Match (\textuparrow), and Mean Dice (\textuparrow).}
\label{tab:combined_results}
                 \resizebox{0.48\textwidth}{!}{%
\begin{tabular}{lccccc}
\hline
\multicolumn{6}{c}{\textbf{LIDC–IDRI}} \\
\hline
Method & GED$\downarrow$ & Dice Soft$\uparrow$ & Dice Max$\uparrow$ & Dice Match$\uparrow$ & Mean Dice$\uparrow$ \\
\hline
U-Net (A1) & 0.306 & 86.59 &  &  & 85.36 \\
U-Net (A2) & 0.246 & 88.43 &  &  & 87.50 \\
U-Net (A3) & 0.244 & 88.20 &  &  & 87.59 \\
U-Net (A4) & 0.296 & 85.83 &  &  & 85.85 \\
CM-Global~\cite{tanno2019learning} & 0.243 & 88.53 & 87.51 & 87.51 & 87.51 \\
CM-Pixel~\cite{zhang2020disentangling} & 0.241 & 88.64 & 87.72 & 87.72 & 87.51 \\
TAB~\cite{liao2023transformer} & 0.232 & 86.35 & 87.11 & 86.06 & 85.97 \\
Pionono~\cite{schmidt2023probabilistic} & 0.150 & 90.00 & 90.10 & 88.97 & 88.84 \\
DPersona~\cite{wu2024diversified} & 0.144 & 90.31 & 90.38 & 89.17 & 89.17 \\
\textbf{Pro\,Sona (ours)} & \textbf{0.120} & \textbf{91.56} & \textbf{92.29} & \textbf{90.26} & \textbf{90.26} \\
\hline
\multicolumn{6}{c}{\textbf{Prostate MRI}} \\
\hline
Method & GED$\downarrow$ & Dice Soft$\uparrow$ & Dice Max$\uparrow$ & Dice Match$\uparrow$ & Mean Dice$\uparrow$ \\
\hline
DPersona~\cite{wu2024diversified} & 0.158 & 87.70 & 90.01 & 86.96 & 86.96 \\
\textbf{Pro\,Sona (ours)} & \textbf{0.146} & \textbf{87.74} & \textbf{90.26} & \textbf{87.02} & \textbf{87.03} \\
\hline
\end{tabular}%
}
\end{table}

\begin{figure}[t]
\centering
\includegraphics[width=\linewidth]{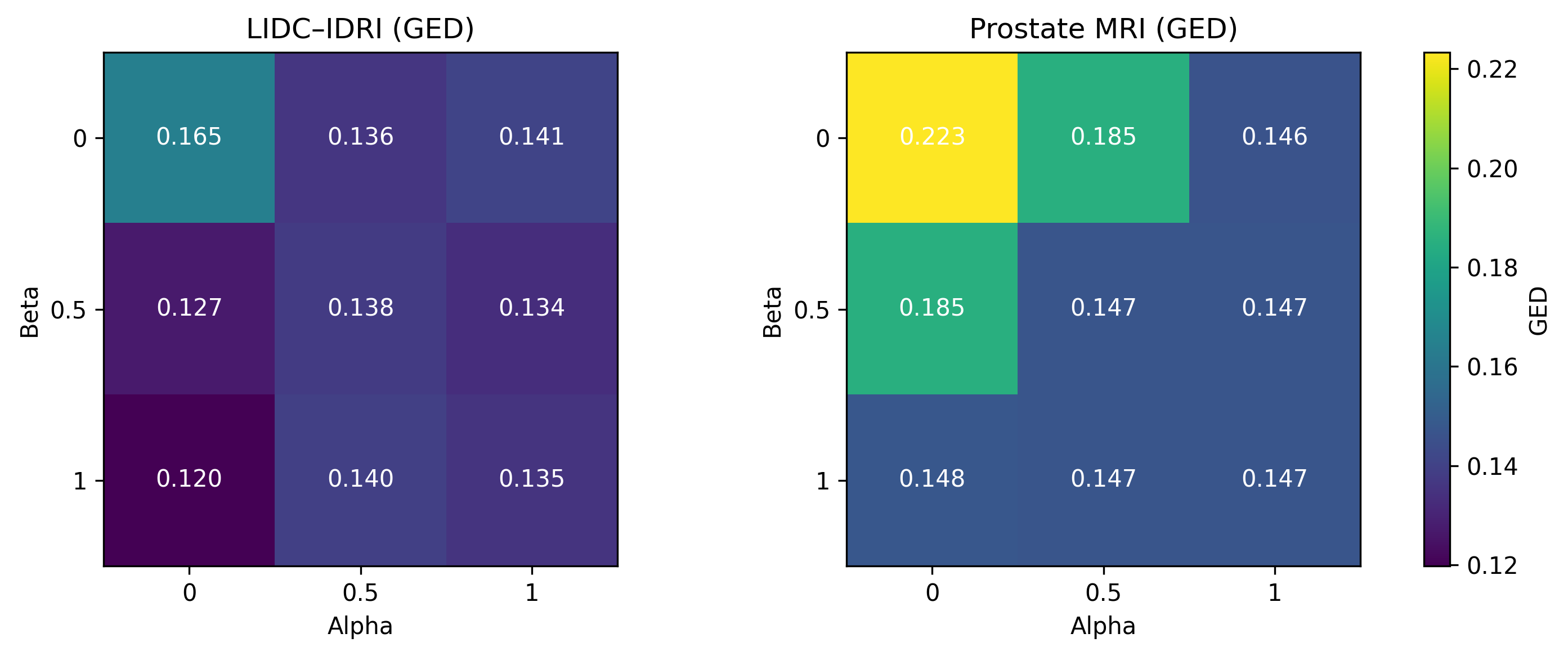}
\caption{Ablation study on the contrastive hyper‑parameters \(\alpha\) and \(\beta\).}
\label{fig:ablation}
\end{figure}

Table~\ref{tab:combined_results} summarises quantitative performance.  
On LIDC–IDRI, \verb|ProSona| achieves the lowest GED and highest Dice metrics across all baselines, reducing GED by 17\% and increasing mean Dice by roughly one point over DPersona.  
On the prostate MRI dataset, \verb|ProSona| consistently surpasses DPersona despite cross-modality and institutional variability, confirming the generalizability of prompt-guided personalization. To assess the sensitivity of Stage~2 to the contrastive weights \(\alpha\) and \(\beta\), we varied them over \(\{0,0.5,1\}\) and tracked GED.
The resulting heatmap in Fig.~\ref{fig:ablation} shows that increasing the similarity‑level weight \(\beta\) consistently lowers GED on LIDC–IDRI, whereas performance on the prostate MRI dataset remains relatively robust across these settings.

\paragraph{Qualitative Analysis.}

\begin{figure*}[t!]
\centering
\includegraphics[width=\linewidth]{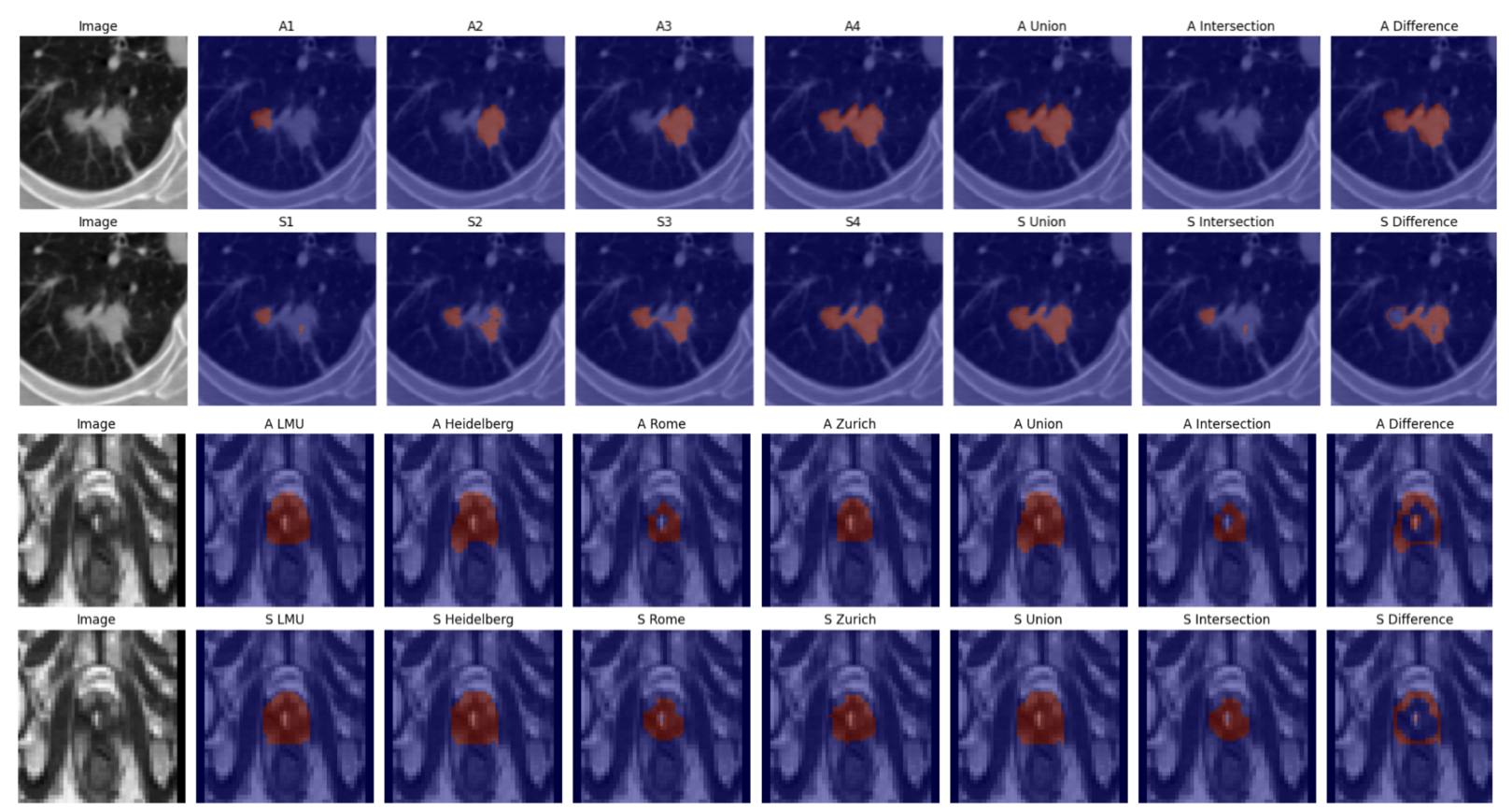}
\caption{Qualitative examples of ProSona on LIDC–IDRI (top) and prostate MRI (bottom). Top: input CT slice with expert annotations (A1,A2,...) and their union/intersection regions. Bottom: predictions (S1,S2,...) guided by prompts describing annotator styles.} %Differences highlight Pro\,Sona’s ability to reflect expert variability.}
\label{fig:qualitative}
\end{figure*}

Figure~\ref{fig:qualitative} shows prompt-conditioned segmentations for both datasets: given prompts such as ``conservative radiologist'' or ``inclusive radiologist,'' \verb|ProSona| reproduces subtle boundary differences matching expert tendencies.  

\begin{figure}[!t]
\centering
\includegraphics[width=0.85\linewidth]{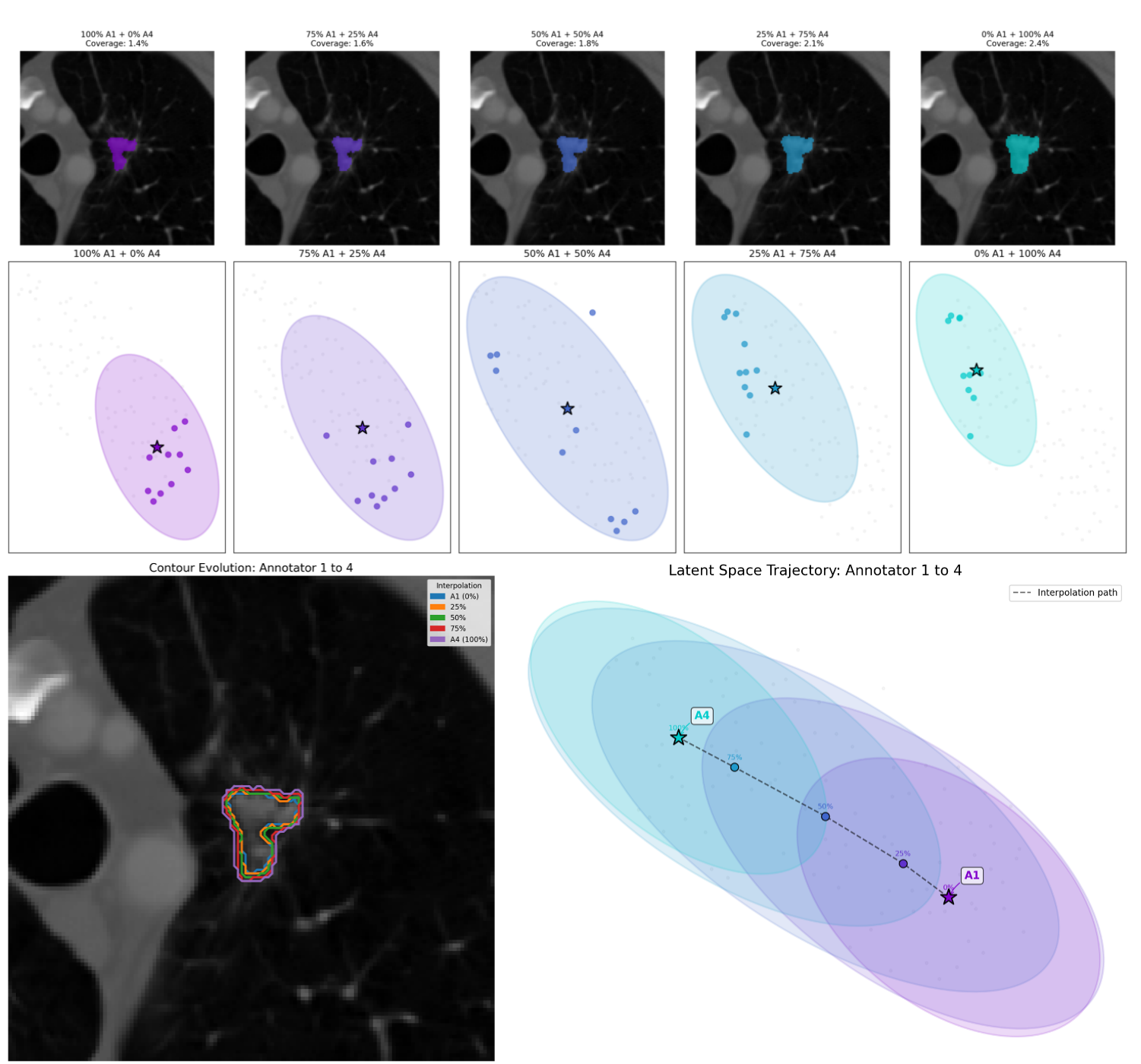}
\caption{Smooth interpolation between annotator styles.}  %As the prompt transitions from a conservative to an inclusive description, ProSona produces segmentations, accordingly.}
\label{fig:interpolation}
\end{figure}

Figure~\ref{fig:interpolation} further demonstrates smooth interpolation between annotator styles, transitioning the prompt from ``small nodule only'' to ``include subtle regions'' causes the segmentation to expand progressively, showcasing the model’s continuous and interpretable control capability, unattainable by discrete multi-head baselines.

% Conclusion 
\section{Discussion and Conclusion}
\label{sec:discussion}

\verb|ProSona| addresses a long-standing limitation of multi-rater segmentation: the inability to model, control, and interpret inter-annotator variability within a unified framework.  
Our two-stage design, comprising a probabilistic latent space and prompt-guided personalization, allows the model to represent multiple plausible expert hypotheses while enabling smooth, meaningful transitions between them.  
Quantitative experiments on LIDC–IDRI and a multi-institutional prostate MRI dataset demonstrate that \verb|ProSona| achieves lower GED and higher Dice scores than prior methods such as DPersona, validating its ability to capture expert annotations.

Our ablation section substantiates the contribution of each component.  
The latent-diversity constraint preserves boundary variability and prevents mode collapse, the prompt-guided mechanism provides direct and interpretable control over segmentation style, and the multi-level contrastive objective aligns textual and visual embeddings to maintain consistent annotator semantics across images.  

Such prompt-guided controllability could facilitate transparent model behavior during contour review in clinical workflows.

Future work will explore richer prompt formulations, extension to 3D volumetric data, and joint modeling of visual–textual uncertainty for active annotation guidance.  
Overall, \verb|ProSona| represents a step toward personalized and interpretable medical image segmentation, bridging the gap between expert diversity and explainability.

\clearpage

\noindent \textbf{Compliance with ethical standards.} 

% Regarding the prostate cancer patients from Heidelberg, all patients were treated within the prospective SMILE trial (NCT04845503): \textit{Stereotactic MRI-guided radiation therapy for localized prostate cancer (SMILE): a prospective, multicentric phase-II-trial}. This study was performed in line with the principles of the Declaration of Helsinki. Approval was granted by the Ethics Committee of Heidelberg University Hospital.

% Informed written consent was obtained from all patients in the scope of ethically approved study protocols in place at the respective institutes: Department of Radiation Oncology of the LMU Munich University Hospital ethics project number 20-291, Heidelberg University Hospital ethics project number S-915/2020, Zurich ethics project number 2021-D0032 and Rome EC authorization number 3460

All procedures involving human participants were conducted in accordance with the ethical standards of the institutional and/or national research committees and with the 1964 Declaration of Helsinki and its later amendments. 

Written informed consent was obtained from all participants under ethically approved study protocols at the respective institutions: Department of Radiation Oncology, LMU Munich University Hospital (project no.\ 20-291); Heidelberg University Hospital (project no.\ S-915/2020); University Hospital Zurich (project no.\ 2021-D0032); and Gemelli University Hospital, Rome (EC authorization no.\ 3460).

For the prostate cancer cohorts from Heidelberg and Zurich, all patients were treated within the prospective SMILE trial (NCT04845503):  \textit{Stereotactic MRI-guided Radiation Therapy for Localized Prostate Cancer (SMILE): A Prospective, Multicentric Phase-II Trial}.

% IEEE-ISBI supports the standard requirements on the use of animal and
% human subjects for scientific and biomedical research. For all IEEE
% ISBI papers reporting data from studies involving human and/or
% animal subjects, formal review and approval, or formal review and
% waiver, by an appropriate institutional review board or ethics
% committee is required and should be stated in the papers. For those
% investigators whose Institutions do not have formal ethics review
% committees, the principles  outlined in the Helsinki Declaration of
% 1975, as revised in 2000, should be followed.

% Reporting on compliance with ethical standards is required
% (irrespective of whether ethical approval was needed for the study) in
% the paper. Authors are responsible for correctness of the statements
% provided in the manuscript. Examples of appropriate statements
% include:
% \begin{itemize}
%   \item ``This is a numerical simulation study for which no ethical
%     approval was required.'' 
%   \item ``This research study was conducted retrospectively using
%     human subject data made available in open access by (Source
%     information). Ethical approval was not required as confirmed by
%     the license attached with the open access data.''
%     \item ``This study was performed in line with the principles of
%       the Declaration of Helsinki. Approval was granted by the Ethics
%       Committee of University B (Date.../No. ...).''
% \end{itemize}

\section*{Acknowledgments}
\label{sec:acknowledgments}
C.K., G.L., and S.A. would like to acknowledge the financial support of the Deutsche Forschungsgemeinschaft (DFG, German Research Foundation) under grant number 469106425 as part of the project “Development of a deep learning toolkit for MRI-guided online adaptive radiotherapy.”
S.A. would also like to acknowledge the additional financial support of the Arab-German Young Academy of Sciences and Humanities (AGYA), which is funded by the German Federal Ministry of Education and Research (BMBF) under grant 01DL20003.
% 

% IEEE-ISBI supports the disclosure of financial support for the project
% as well as any financial and personal relationships of the author that
% could create even the appearance of bias in the published work. The
% authors must disclose any agency or individual that provided financial
% support for the work as well as any personal or financial or
% employment relationship between any author and the sources of
% financial support for the work.

% Other types of acknowledgements can also be listed in this section.

% Reporting on real or potential conflicts of interests, or the absence
% thereof, is required in the paper. Authors are responsible for
% correctness of the statements provided in the manuscript. Examples of
% appropriate statements include:
% \begin{itemize}
%   \item ``No funding was received for conducting this study. The
%     authors have no relevant financial or non-financial interests to
%     disclose.'' 
%   \item ``This work was supported by […] (Grant numbers) and
%     […]. Author X has served on advisory boards for Company Y.'' 
%   \item ``Author X is partially funded by Y. Author Z is a Founder and
%     Director for Company C.''
% \end{itemize}

% -------------------------------------------------------------------------
% References appear on a separate page.
\bibliographystyle{IEEEbib}
\bibliography{refs_2ds}

\end{document}